\relax
\documentclass[letterpaper]{article} 
\usepackage[accepted]{icmlstyle} 
\usepackage{times}  
\usepackage{helvet}  
\usepackage{courier}  
\usepackage{url}  
\usepackage[T1]{fontenc}
\usepackage{graphicx}  
\usepackage{amsmath}
\usepackage{mathrsfs}
\usepackage{nicefrac}
\usepackage{amsfonts}
\usepackage{combelow}
\usepackage[dvipsnames]{xcolor}
\usepackage{booktabs}
\usepackage{multirow}
\usepackage{adjustbox}
\usepackage{accents}
\usepackage{subfig}
\usepackage{array}
\usepackage{enumitem}
\usepackage{hyperref}
\usepackage{booktabs} 
\usepackage{bm}
\usepackage{tipa}
\usepackage[small,compact]{titlesec}
\usepackage{combelow}
\newcommand{\topic}[1]{\vspace{0.01in}\noindent\textbf{#1.}} 
\newcommand{\nstopic}[1]{\noindent\textbf{#1.}} 

\newcommand{\PreserveBackslash}[1]{\let\temp=\\#1\let\\=\temp}
\newcolumntype{C}[1]{>{\PreserveBackslash\centering}p{#1}}
\newcolumntype{R}[1]{>{\PreserveBackslash\raggedleft}p{#1}}
\newcolumntype{L}[1]{>{\PreserveBackslash\raggedright}p{#1}}

\newcommand{\splitcell}[4]
{
	\begin{tabular}{@{}R{1.3cm}|L{1.3cm}@{}}
		#1 & #2 \\
		#3 & #4 \\
	\end{tabular}
}

\newcommand{\splitcellc}[4]
{
	\begin{tabular}{@{}R{1.3cm}|L{1.3cm}@{}}
		#1 & #2 \\
		#3 & #4 \\
	\end{tabular}
}

\newcommand{\splitcellr}[4]
{
	\begin{tabular}{@{}R{1.3cm}|L{1.3cm}@{}}
		#1 & #2 \\
		#3 & #4 \\
	\end{tabular}
}

\newcolumntype{?}[1]{!{\ width #1}}

\begin{document}

\twocolumn[
\icmltitle{On the Effectiveness of Regularization Against Membership Inference Attacks}



\icmlsetsymbol{equal}{*}

\begin{icmlauthorlist}
\icmlauthor{Yigitcan Kaya}{to}
\icmlauthor{Sanghyun Hong}{to}
\icmlauthor{Tudor Dumitra\cb{s}}{to}

\end{icmlauthorlist}

\icmlaffiliation{to}{University of Maryland, Maryland, USA}

\icmlcorrespondingauthor{Yigitcan Kaya}{cankaya@umiacs.umd.edu}

\icmlkeywords{Machine Learning, Deep Learning, Privacy, Adversarial Machine Learning}

\vskip 0.3in
]



\printAffiliationsAndNotice{}  

%
%

\begin{abstract}
Deep learning models often raise privacy concerns as they leak information about their training data. 
This enables an adversary to determine whether a data point was in a model's training set by conducting a membership inference attack (MIA). 
Prior work has conjectured that regularization techniques, which combat overfitting, may also mitigate the leakage. 
While many regularization mechanisms exist, their effectiveness against MIAs has not been studied systematically, and the resulting privacy properties are not well understood. 
We explore the lower bound for information leakage that practical attacks can achieve. 
First, we evaluate the effectiveness of 8 mechanisms in mitigating two recent MIAs, on three standard image classification tasks. 
We find that certain mechanisms, such as label smoothing, may inadvertently help MIAs.  
Second, we investigate the potential of improving the resilience to MIAs by combining complementary mechanisms.
Finally, we quantify the opportunity of future MIAs to compromise privacy by designing a white-box \emph{distance-to-confident} (DtC) metric, based on adversarial sample crafting. 
Our metric reveals that, even when existing MIAs fail, the training samples may remain distinguishable from test samples. 
This suggests that regularization mechanisms can provide a false sense of privacy, even when they appear effective against existing MIAs.
\end{abstract}
%
%

\section{Introduction}
\label{sec:intro}
Deep learning has emerged as one of the cornerstones of large-scale machine learning.
However, its fundamental dependence on data forces practitioners to collect and use personal information~\cite{shokri2015privacy}.
This situation has given rise to privacy concerns as deep learning models are shown to leak information about their data through their behaviors or structures~\cite{fredrikson2015model,yang2019adversarial,carlini2018secret,shokri2017membership}.
In particular, membership inference attacks (MIAs) enable an adversary to find out whether a data point was in a model's training set.
This often represents a serious privacy threat; for example, learning that an individual is in a hospital's diagnosis training data via an MIA also reveals that this individual was a patient there.
%

MIAs are possible when the adversary can distinguish between a model's predictions on training set and test set samples. 
In this case, information that leaks from the model allows the adversary to infer which samples were used for training the model. 
Differential privacy (DP) provides a formal upper bound for this information leakage~\cite{abadi2016deep}.
While this guarantees that the model is not vulnerable to MIAs, prior work suggests that DP is hard to apply without hurting the model's utility~\cite{abadi2016deep,rahman2018membership}, and its guarantees might be too restrictively conservative against known attacks~\cite{jayaraman2019evaluating}.
%
%
As utility is a key factor for the adoption of security mechanisms in practice, researchers have also made a connection between the vulnerability to MIAs and \emph{overfitting}, which increases the model's prediction confidence for the samples in the training set. 
In consequence, prior work has conjectured that regularization may represent a practical defense against MIAs, as it discourages models from overfitting to their training data~\cite{yeom2017unintended,shokri2017membership,salem2018ml}.
%

The prior work leaves several open questions. 
While many regularization mechanisms have been proposed over the past few years~\cite{gastaldi2017shake,wan2013regularization,icml19shallowdeepnetworks,hinton2015distilling}, it is unclear whether all mechanisms are effective, or which are more effective against MIAs.
Furthermore, it is unknown whether they prevent the \emph{same leakage}, or their effects can complement each other for better resilience to MIAs. 
More importantly, the effectiveness of regularization mechanisms against existing MIAs may not reflect the vulnerability of models to future attacks.

In this paper, we systematically analyze the 
effectiveness of regularization mechanisms against MIAs.
Unlike the \emph{theoretical upper bound for leakage} provided by differential privacy, we explore the \emph{lower bound} that practical attacks can achieve.
We first evaluate 8 popular regularization mechanisms against two state-of-the-art MIAs using modern convolutional neural networks, on three popular image classification tasks: Fashion-MNIST, CIFAR-10, and CIFAR-100.
We find that 
(i) some mechanisms, such as distillation and random data augmentation, are more effective than others in mitigating MIAs, including DP; 
(ii) applying regularization for only increasing a model's utility leads to less-than-ideal protection against MIAs; and 
(iii) some mechanisms, especially label smoothing~\cite{szegedy2016rethinking}, might significantly exacerbate the vulnerability over a standard model.
Building on our findings, we propose guidelines on selecting and applying regularization against MIAs for practitioners.
We also see that increasing the intensity of a single mechanism, e.g., distillation, might bring diminishing returns for privacy.
This leads us to ask whether it is beneficial to combine multiple mechanisms.

Our evaluation considers each mechanism independently; however, it might be possible to apply them in conjunction.
Prior work suggests that different regularization mechanisms alter the model in distinct ways.
For example, dropout is thought to approximate a large ensemble of networks~\cite{baldi2013understanding}; whereas label smoothing changes the geometry of the activations and forces well-separated clusters~\cite{muller2019does}.
Given these diverse effects, we also investigate the benefits of combining multiple regularization mechanisms against MIAs.
We observe that combining an effective mechanism with a less effective one might improve upon only applying the effective mechanism by itself.
To mitigate MIAs, we believe that finding optimal combinations is an important research challenge.

The success against known MIAs provides a lower bound for the privacy leakage vulnerability.
Against a potentially stronger attack, however, this may give a false sense of privacy.
In the absence of formal guarantees, we aim to provide more realistic lower bounds by designing a white-box metric, \emph{distance-to-confident} (DtC), based on an adversarial example crafting algorithm~\cite{madry2017towards}.
Specifically, we slowly perturb a data sample until the model classifies it with high confidence, and our metric quantifies the amount of perturbation required. 
Intuitively, if a model has no leakage, the DtC scores on its training and test samples would be similar.
However, even in cases when existing MIAs fail completely, the DtC gap between training and test samples still persists. 
This discrepancy may be exploited by an unknown future attack to tell the training samples apart.
When we apply our metric to differentially-private models, the DtC gap disappears as the guarantees get stronger.
This shows that stronger DP guarantees are actually useful against MIAs and not as overly conservative as previously thought.

In summary, we make the following contributions:

\textbullet We systematically evaluate 8 popular regularization mechanisms against existing membership inference attacks (MIAs). We identify the pitfalls and the best practices of using regularization to defeat MIAs (Section~\ref{sec:effectiveness}). \\
\textbullet We investigate the benefits of combining multiple mechanisms against MIAs and find the combinations that provide additional resilience with less utility damage (Section~\ref{sec:combining}).  \\
\textbullet We design the \emph{distance-to-confident} (DtC) metric based on adversarial perturbations and reveal that mechanisms that eliminate MIAs might still result in discrepancies between the training and the test sets. This hints that resilience against existing MIAs might give a false sense of privacy (Section~\ref{sec:comparison}).

%
%

\section{Related Work}

\nstopic{Membership Inference Attacks on Deep Learning}
Membership inference attacks (MIAs) have been proposed as an indicator of fundamental privacy flaws in deep learning.
\citet{shokri2017membership} propose an MIA based on training an inference model to distinguish between a model's predictions on training and test set samples.
\citet{yeom2017unintended} propose a simpler, but equally effective, attack that compares the model's prediction loss (or confidence) on a target sample with the model's average loss on its training set.
Finally, building on Shokri et al.'s work, \citet{salem2018ml} have developed more effective attacks that require weaker assumptions.
These studies have suggested simple regularization techniques as a countermeasure against MIAs. 
The lack of comprehensive evaluation on the effectiveness of regularization, however, prevents it from being considered as a defense, in practice.
We use these attacks to evaluate a wide range of regularization techniques on modern tasks and investigate whether regularization is actually effective.

\topic{Deep Learning with Differential Privacy}
Differential privacy (DP) offers a formal method to measure and limit an algorithm's privacy leakage~\cite{dwork2008differential}.
Building on this framework, \citet{abadi2016deep} have proposed a deep learning training algorithm, differentially private stochastic gradient descent (DP-SGD). 
DP-SGD, by clipping and adding noise to the parameter updates, ensures that a model's leakage stays within a privacy budget, $\varepsilon$.
However, as DP focuses on the worst-case leakage, \citet{jayaraman2019evaluating} have shown that its guarantees might be too strict against realistic attacks.
In the same context, \citet{rahman2018membership} suggest that DP-SGD mitigates existing MIAs at the cost of the model's utility.
We use DP as a baseline to compare against as its resilience against MIAs is backed by formal guarantees.

\topic{Generalization and Memorization in Deep Learning}
Recent work has attempted to explain why deep learning models generalize well to unseen samples, despite their tendency to \emph{memorize}.
\citet{arpit2017closer} and \citet{zhang2016understanding} have shown that a model can memorize random data and labels with almost perfect accuracy, which indicates their capacity to overfit to their training data.
These studies have also shown that regularization techniques, such as dropout~\cite{srivastava2014dropout} or weight decay~\cite{krogh1992simple}, might hinder memorization.
Further, several studies have proposed deep learning regularization techniques for improving generalization, such as label smoothing~\cite{szegedy2016rethinking}, small batch size~\cite{keskar2016large}, and shake-shake~\cite{gastaldi2017shake}.
We investigate the ties between regularization, generalization, and vulnerability of deep learning models to MIAs.

\section{Experimental Setup}

\nstopic{Datasets}
We use three datasets for benchmarking: Fashion-MNIST~\cite{xiao2017fashion}, CIFAR-10 and CIFAR-100~\cite{krizhevsky2009learning}.
The Fashion-MNIST consists of 28x28 pixels, gray-scale images of fashion items drawn from 10 classes; containing 60,000 training and 10,000 validation images.
The CIFAR-10 and CIFAR-100 consist of 32x32 pixels, colored natural images drawn from 10 and 100 classes, respectively; containing 50,000 training and 10,000 validation images.
We train our baseline models without any data augmentation; however, we study random cropping as a regularization mechanism.
These tasks represent different levels of learning complexity and risk of overfitting; Fashion-MNIST being the easiest task and CIFAR-100 being the most difficult as it has a large number of classes and few samples per class.

\topic{Architectures and Hyperparameters}
We experiment with variants of a popular convolutional neural network (CNN) architecture, VGG~\cite{simonyan2014very}.
VGG is a prototypical architecture, on which complex architectures are built.
For Fashion-MNIST, CIFAR-10, and CIFAR-100, we use 4-layer, 7-layer, and 9-layer CNNs, respectively.
We train our models for 30 epochs, with $0.0002$ learning rate, using the ADAM optimizer from~\cite{reddi2019convergence}.
By default, we set the $L_2$ weight decay coefficient to $10^{-6}$ and the batch size to $128$.
We run our experiments several times to compensate for the randomness in training deep learning models.

\topic{Metrics}
To quantify a model's utility, we use its top-1 accuracy on the validation set, $Acc$, as a percentage.
An MIA's accuracy, $Inf$, is the probability that the adversary can guess correctly whether an input is from the training set or not.
We apply the MIAs on a set of data points, sampled from the training and testing sets of the model with an equal probability.
Note that, on this data set, a random guessing strategy will lead to 50\% inference accuracy.
To quantify the success of an MIA, we use the adversary advantage metric~\cite{yeom2017unintended} as a percentage, i.e., $Adv = 2 \times (Inf - 50\%)$.
Finally, to quantify a mechanism's impact on utility, we measure the relative accuracy drop (RAD) over a baseline model, i.e., $\nicefrac{(Acc_{baseline} - Acc_{mechanism})}{Acc_{baseline}}$.

\section{Effectiveness of Regularization}
\label{sec:effectiveness}

\topic{Setting}
We consider the supervised learning setting and the standard feed-forward deep neural network (DNN) structure for classification.
A DNN model, a classifier, is a function, $F$, that maps an input sample $x$, e.g., a natural image, to an output vector $\hat{y}$, e.g., the probabilities of image $x$ belonging to each class label.
The model then classifies $x$ in the most likely class, i.e., $\hat{y}_i = \operatorname*{argmax}_i \hat{y}$.
The parameters of a DNN are learned on a private training set, $\mathcal{S}$, containing multiple $(x,y)$ pairs; where $y$ is the ground-truth label of $x$.
During training, the model's parameters are updated to minimize its \emph{loss} $\mathcal{L}(\hat{y}, y)$, i.e., a measure of how off $\hat{y}$ is from $y$, on the samples in $\mathcal{S}$.
After training, the model is tested on previously unseen data samples, $\mathcal{D}$.
Because of their immense learning capacity, modern DNNs usually \emph{overfit} to their training set and have a large \emph{generalization gap}, i.e., their performance on $\mathcal{S}$ diverges from their performance on $\mathcal{D}$.
As a result of overfitting, $F$ produces more accurate and more confident predictions on $\mathcal{S}$ than on $\mathcal{D}$.

\topic{Attacks}
Membership inference attacks (MIAs) aim to find out whether a particular sample, $(x_t, y_t)$, was in $\mathcal{S}$ of the victim model, $F_v$.
They exploit the generalization gap and the discrepancies between $F_v$'s predictions on $\mathcal{S}$ and $\mathcal{D}$.
As we are investigating ways to mitigate this vulnerability, we consider a strong adversary that has (i) query access to $F_v$, i.e., $F_v(x) \forall x$; (ii) the average loss, $\overline{\mathcal{L}}$, of $F_v$ on the samples in $\mathcal{S}$; and (iii) some input samples from both $\mathcal{D}$ and $\mathcal{S}$ of the $F_v$.
In our experiments, we use two state-of-the-art MIAs from \citet{yeom2017unintended} and \citet{salem2018ml}.
In Yeom et al.'s attack, the adversary obtains $F_v(x_t) = \hat{y}_t$ and computes $\mathcal{L}_t = \mathcal{L}(\hat{y_t}, y_t)$. 
The adversary then infers that $x_t$ is in $\mathcal{S}$ if $\mathcal{L}_t < \overline{\mathcal{L}}$.
In Salem et al.'s attack, the adversary collects $F_v(x_i) \forall x_i \in \mathcal{A}$, where $\mathcal{A}$ is a set of samples from $\mathcal{D}$ and $\mathcal{S}$.
Then the adversary, using $\mathcal{A}$ as its training set, learns a binary classifier to classify $F_v(x_t)$ as either being in $\mathcal{S}$ or not.

\topic{Regularization Mechanisms}
Regularization is any modification to a learning algorithm that aims to reduce its generalization gap and to mitigate overfitting.
We evaluate 8 common mechanisms that alter different parts of deep learning pipeline: (i) training label transformations; (ii) training data perturbations; (iii) architectural modifications; and (iv) optimization methods.
In (i), we evaluate distillation with temperature~\cite{hinton2015distilling} and label smoothing~\cite{szegedy2016rethinking}. 
In (ii), we evaluate adding Gaussian noise to training samples~\cite{arpit2017closer} and data augmentation via padding then randomly cropping~\cite{simonyan2014very}.
In (iii), we evaluate dropout after convolutional layers~\cite{szegedy2016rethinking} and spatial dropout~\cite{tompson2015efficient}.
In (iv), we evaluate weight decay~\cite{krogh1992simple} and early stopping at an epoch~\cite{caruana2001overfitting}.
Against mechanisms that manipulate the labels or the data, we assume that the adversary only knows the original distribution.
We give hyperparameter ranges for each mechanism in Table~\ref{table:reg_mechanisms}.

\begin{table}[htb]
	 \caption{The regularization mechanisms and their associated hyperparameters.}
    \label{table:reg_mechanisms}
    \centering
    \begin{small}
    \begin{sc}
    \begin{tabular}{lc}
        \toprule
        \textbf{Mechanism} & \textbf{Hyperparameter Range}\\
        \midrule
        \small{Distillation ($T$)}&\small{$1 \leq T \leq 100$}\\
        \small{Label Smoothing ($\alpha$)}&\small{$1 \leq \alpha \leq 0.99$}\\
        \small{Gaussian Noise ($\sigma$)}&\small{$0.01 \leq \sigma \leq 0.25$}\\
        \small{Random Cropping ($c$)}&\small{$1 \leq c \leq 24$}\\
        \small{Dropout} ($p$) &\small{$0.05 \leq p \leq 0.95$}\\
        \small{Spat. Dropout ($q$)}&\small{$0.05 \leq q \leq 0.95$}\\
        \small{Weight Decay ($\lambda$)}&\small{$1e-6 \leq \lambda \leq 0.5$}\\
        \small{Early Stopping ($E$)}&\small{$30 \geq E \geq 1$}\\
        \bottomrule
    \end{tabular}
    \end{sc}
    \end{small}
\end{table}
\raggedbottom

\subsection{Finding the Most Effective Mechanisms}
\label{ssec:effectiveness}
To use as our baselines, we present our CNNs with no regularization in Table~\ref{table:baseline}.
For each task's baseline model, the table includes its accuracy, $Acc$, and its adversary advantage rates against MIAs, $Adv_{Yeom}$, and $Adv_{Salem}$.
Note that MIAs are less threatening on the simpler task, Fashion-MNIST, as the model already achieves high accuracy on unseen samples.
This highlights that the risk of MIAs is even greater for modern, large-scale tasks and further motivates our study.
Moreover, we also see that Yeom et al.'s MIA usually leads to a higher adversary advantage than Salem et al.'s.
Thus, in our experiments, we use Yeom et al.'s MIA to identify the most effective configurations.

\begin{table}[htb]
    \caption{The baseline models with no regularization. We present their validation accuracies and the membership inference advantage rates against them.}
    \label{table:baseline}
    \centering
    \begin{small}
    \begin{sc}
    \begin{tabular}{lccr}
        \toprule
        \textbf{Task} &\textbf{$Acc$} &\textbf{$Adv_{Yeom}$} &\textbf{$Adv_{Salem}$}\\
        \midrule
        \small{Fashion-MNIST}&\small{92.6\%}&\small{9.6\%}&\small{8.3\%}\\
        \small{CIFAR-10}&\small{81.9\%}&\small{35.8\%}&\small{18.5\%}\\
        \small{CIFAR-100}&\small{57.3\%}&\small{68.9\%}&\small{49.9\%}\\
        \bottomrule
    \end{tabular}
    \end{sc}
    \end{small}
\end{table}
\raggedbottom

As the mechanisms we apply might hurt a model's accuracy, we evaluate our results under three different utility scenarios: maximum utility ($Max$), $<$5\% and $<$15\% relative accuracy drop (RAD) over the corresponding baseline model. 
Our scenarios focus on having reasonable utility for practitioners, based on the standards prior work sets~\cite{hong2019terminal}.
In the $Max$ scenario, to give an idea about how MIAs fare when we apply regularization only for increasing utility, we find the setting of each mechanism that results in the highest accuracy.
In the RAD scenarios, we find the settings that yield models with the smallest adversary advantage, within the respective RAD limit.

We present the results on Fashion-MNIST, CIFAR-10, and CIFAR-100 in Tables~\ref{table:fmnist_results},~\ref{table:cifar10_results} and~\ref{table:cifar100_results}, respectively.
We first see that increasing regularization intensity usually leads to more accuracy drop and, also, decreases the MIA success, by hindering the model's ability to overfit.
Further, mitigating MIAs is significantly easier on Fashion-MNIST than on more complex tasks.
On Fashion MNIST, within 5\% RAD, regularization reduces $Adv_{Yeom}$, on average, by 82\% and up to 95\%.
On CIFAR-10, after regularization $Adv_{Yeom}$ drops, on average, by 44\%, and up to 93\%; and on CIFAR-100, it drops by 35\% and up to 80\%.
Within 15\% RAD, on Fashion-MNIST, regularization reduces $Adv_{Yeom}$, on average, by 91\% and up to 100\%; on CIFAR-10, by 70\% and, up to, 94\%; and on CIFAR-100, by 41\% and, up to, 87\%.
Our results show that eliminating MIAs on complex tasks is challenging even after incurring significant utility losses.

Note that applying regularization for maximum accuracy does not lead to a significant reduction in MIA success, except for random cropping and distillation mechanisms.
In the $Max$ scenario, regularization reduces $Adv_{Yeom}$, on average, by only 40\%, 22\%, and 14\%; on Fashion-MNIST, CIFAR-10, and CIFAR-100, respectively. 
Further, decreasing the generalization gap and increasing $Acc$ via regularization usually reduces MIA success; however, in certain cases, such as label smoothing, it can have the opposite effect.
This suggests that the connection between the generalization gap and the vulnerability to MIAs might not be as clear as previously thought~\cite{yeom2017unintended}.


\begin{table}[htb]
	\caption{Applying regularization mechanisms on Fashion-MNIST, in three utility scenarios. Each cell contains the mechanism's hyperparameter and the relative changes over the baseline model in $Acc$, $Adv_{Yeom}$, and $Adv_{Salem}$; in its top left, top right, bottom left, and bottom right quadrants, respectively. In terms of $Adv$, `*' marks the highest drop  and `\textdagger' marks an inadvertent increase.}
	\label{table:fmnist_results}
	\centering
	\adjustbox{max width=\linewidth}{%
		\begin{tabular}{@{}c@{}c@{}c@{}}
			\toprule
			\textbf{MAX}  & \textbf{RAD < 5\%}  & \textbf{RAD < 15\%}\\
			\midrule
			\splitcell{$T\!=\!2$}{+0.1\%}{-33.3\%}{-9.6\%}  &  \splitcellc{$T\!=\!50$}{-2.9\%}{-94.8\%*}{-100\%*}  &  \splitcellr{$T\!=\!50$}{-2.9\%}{-94.8\%}{-100\%*}\\ \addlinespace[0.14cm]
			
			\splitcell{$\alpha\!=\!.01$}{+0.2\%}{-8.3\%}{+34.9\%\textdagger} & \splitcellc{$\alpha\!=\!.9$}{-1.5\%}{-63.5\%}{-100\%*} & \splitcellr{$\alpha\!=\!.99$}{-10.8\%}{-95.8\%}{-100\%*}\\ \addlinespace[0.14cm]
			
			\splitcell{$\sigma\!=\!.025$}{-0.9\%}{-66.7\%}{-73.5\%} & \splitcellc{$\sigma\!=\!.2$}{-4.1\%}{-81.3\%}{-100\%*} & \splitcellr{$\sigma\!=\!.225$}{-6.7\%}{-84.4\%}{-100\%*}\\ \addlinespace[0.14cm]
			
			\splitcell{$c\!=\!1$}{-0.1\%}{-81.3\%*}{-81.9\%*} & \splitcellc{$c\!=\!4$}{-3.0\%}{-90.7\%}{-100\%*} & \splitcellr{$c\!=\!18$}{-9.9\%}{-100\%*}{-100\%*}\\ \addlinespace[0.14cm]
			
			\splitcell{$p\!=\!.5$}{+0.2\%}{-55.2\%}{-73.5\%} & \splitcellc{$p\!=\!.95$}{-2.5\%}{-85.4\%}{-100\%*} & \splitcellr{$p\!=\!.95$}{-2.5\%}{-85.4\%}{-100\%*}\\ \addlinespace[0.14cm]
			
			\splitcell{$q\!=\!.25$}{+0.2\%}{-52.1\%}{-100\%} & \splitcellc{$q\!=\!.75$}{-2.6\%}{-87.5\%}{-100\%*} & \splitcellr{$q\!=\!.95$}{-11.0\%}{-89.6\%}{-100\%*}\\ \addlinespace[0.14cm]
			
			\splitcell{$\lambda\!=\!1e\!-\!4$}{+0.1\%}{-1.0\%}{+9.6\%\textdagger} & \splitcellc{$\lambda\!=\!.01$}{-1.5\%}{-72.9\%}{-97.6\%} & \splitcellr{$\lambda\!=\!.1$}{-9.1\%}{-93.8\%}{-100\%*}\\ \addlinespace[0.14cm]
			
			\splitcell{$E\!=\!20$}{+0.1\%}{-28.1\%}{-19.3\%} & \splitcellc{$E\!=\!1$}{-4.9\%}{-80.2\%}{-96.4\%} & \splitcellr{$E\!=\!1$}{-4.9\%}{-80.2\%}{-96.4\%}\\
			\bottomrule
		\end{tabular}
	}
\end{table}
\raggedbottom


\begin{table}[t]
	\caption{Applying regularization mechanisms on CIFAR-10, in three utility scenarios.}
	\label{table:cifar10_results}
	\centering
	\adjustbox{max width=\linewidth}{%
		\begin{tabular}{@{}c@{}c@{}c@{}}
			\toprule
			\textbf{MAX}  & \textbf{RAD < 5\%}  & \textbf{RAD < 15\%}\\
			\midrule
			\splitcell{$T\!=\!5$}{+1.3\%}{-50.3\%}{+13.5\%\textdagger} & \splitcellc{$T\!=\!25$}{-2.4\%}{-78.2\%}{-100\%*} & \splitcellr{$T\!=\!50$}{-8.2\%}{-84.7\%}{-100\%*}\\ \addlinespace[0.14cm]
			
			\splitcell{$\alpha\!=\!.01$}{+1.2\%}{-31.0\%}{+33.5\%\textdagger} & \splitcellc{$\alpha\!=\!.01$}{+1.2\%}{-31.0\%}{+33.5\%\textdagger} & \splitcellr{$\alpha\!=\!.99$}{-10.1\%}{-46.1\%}{-100\%*}\\ \addlinespace[0.14cm]
			
			\splitcell{$\sigma\!=\!.01$}{-1.0\%}{-0.1\%}{+9.2\%\textdagger} & \splitcellc{$\sigma\!=\!.025$}{-4.2\%}{-21.5\%}{-18.4\%} & \splitcellr{$\sigma\!=\!.05$}{-11.1\%}{-44.9\%}{-42.2\%}\\ \addlinespace[0.14cm]
			
			\splitcell{$c\!=\!2$}{+3.5\%}{-48.6\%}{-31.9\%} & \splitcellc{$c\!=\!22$}{-3.7\%}{-92.7\%*}{-72.8\%} & \splitcellr{$c\!=\!24$}{-7.7\%}{-93.8\%*}{-78.4\%}\\ \addlinespace[0.14cm]
			
			\splitcell{$p\!=\!.25$}{-1.9\%}{-8.7\%}{+3.2\%\textdagger} & \splitcellc{$p\!=\!.5$}{-3.4\%}{-28.2\%}{-16.8\%} & \splitcellr{$p\!=\!.95$}{-9.3\%}{-63.4\%}{-42.2\%}\\ \addlinespace[0.14cm]
			
			\splitcell{$q\!=\!.1$}{+0.6\%}{-20.4\%}{+23.2\%\textdagger} & \splitcellc{$q\!=\!.25$}{-0.6\%}{-59.5\%}{-36.2\%} & \splitcellr{$q\!=\!.5$}{-12.9\%}{-92.2\%}{-89.2\%}\\ \addlinespace[0.14cm]
			
			\splitcell{$\lambda\!=\!5e\!-\!4$}{+1.0\%}{-17.0\%}{+35.7\%\textdagger} & \splitcellc{$\lambda\!=\!5e\!-\!4$}{+1.0\%}{-17.0\%}{+35.7\%\textdagger}  &  \splitcellr{$\lambda=.1$}{-13.3\%}{-81.3\%}{-92.4\%}\\ \addlinespace[0.14cm]
			
			\splitcell{$E\!=\!30$}{+0.0\%}{+0.0\%}{+0.0\%} &  \splitcellc{$E\!=\!20$}{-3.4\%}{-25.4\%}{-21.1\%} & \splitcellr{$E\!=\!2$}{-13.9\%}{-84.1\%}{-72.4\%}\\ \bottomrule
		\end{tabular}
	}
\end{table}
\raggedbottom


\begin{table}[t]
	\caption{Applying regularization mechanisms on CIFAR-100, in three utility scenarios.}
	\label{table:cifar100_results}
	\centering
	\adjustbox{max width=\linewidth}{%
		\begin{tabular}{@{}c@{}c@{}c@{}}
			\toprule
			\textbf{MAX}  & \textbf{RAD < 5\%}  & \textbf{RAD < 15\%}\\
			\midrule
			\splitcell{$T\!=\!3$}{+3.1\%}{-38.3\%}{+23.8\%\textdagger}  & \splitcellc{$T\!=\!50$}{-4.2\%}{-68.6\%}{-99.4\%*}  &  \splitcellr{$T\!=\!100$}{-6.1\%}{-68.8\%}{-100\%*}\\ \addlinespace[0.14cm]
			
			\splitcell{$\alpha\!=\!.5$}{+5.9\%*}{-32.2\%}{+58.7\%\textdagger}  & \splitcellc{$\alpha\!=\!.95$}{+5.6\%}{-36.4\%}{+23.4\%\textdagger}  & \splitcellr{$\alpha\!=\!.95$}{+5.6\%}{-36.4\%}{+23.4\%\textdagger}\\ \addlinespace[0.14cm]
			
			\splitcell{$\sigma\!=\!.01$}{+0.2\%}{-2.9\%}{-4.4\%} & \splitcellc{$\sigma\!=\!.01$}{+0.2\%}{-2.9\%}{-4.4\%} & \splitcellr{$\sigma\!=\!.025$}{-6.5\%}{-16.7\%}{-16.0\%}\\ \addlinespace[0.14cm]
			
			\splitcell{$c\!=\!2$}{+3.3\%}{-24.7\%}{-19.8\%} & \splitcellc{$c\!=\!16$}{-4.4\%}{-79.8\%*}{-87.6\%}  & \splitcellr{$c\!=\!20$}{-5.9\%}{-86.9\%*}{-100\%*}\\ \addlinespace[0.14cm]
			
			\splitcell{$p\!=\!.1$}{-0.9\%}{-0.2\%}{-8.4\%} & \splitcellc{$p\!=\!.25$}{-1.4\%}{-7.8\%}{-9.4\%} & \splitcellr{$p\!=\!.5$}{-5.9\%}{-17.1\%}{-15.2\%}\\ \addlinespace[0.14cm]
			
			\splitcell{$q\!=\!.05$}{-1.9\%}{-10.6\%}{-10.4\%} & \splitcellc{$q\!=\!.25$}{-2.0\%}{-52.9\%}{-53.1\%} &  \splitcellr{$q\!=\!.25$}{-2.0\%}{-52.9\%}{-53.1\%}\\ \addlinespace[0.14cm]
			
			\splitcell{$\lambda\!=\!1e\!-\!5$}{+1.2\%}{+1.0\%\textdagger}{-6.8\%} & \splitcellc{$\lambda\!=\!5e\!-\!4$}{-2.4\%}{-18.9\%}{+26.6\%\textdagger} & \splitcellr{$\lambda\!=\!5e\!-\!4$}{-2.4\%}{-18.9\%}{+26.6\%\textdagger}\\ \addlinespace[0.14cm]
			
			\splitcell{$E\!=\!30$}{+0.0\%}{+0.0\%}{+0.0\%} & \splitcellc{$E\!=\!15$}{-1.1\%}{-7.7\%}{-1.4\%} & \splitcellr{$E\!=\!10$}{-10.3\%}{-33.1\%}{-31.9\%}\\ \bottomrule
		\end{tabular}
		}
\end{table}
\raggedbottom

\topic{Mechanism Consistency}
As we aim is to provide guidelines for practitioners, we evaluate regularization mechanisms in terms of their consistency across the board.
We first see that distillation and random cropping are the most effective and consistent mechanisms.
Within 5\% RAD, random cropping reduces $Adv_{Yeom}$ and $Adv_{Salem}$ by 87\%; and distillation reduces $Adv_{Yeom}$ by 81\% and $Adv_{Salem}$ by 100\%.
However, considering the small drop in MIA success between 5\% RAD and 15\% RAD scenarios; we observe diminishing returns, especially for distillation.

Prior work suggests dropout and weight decay against MIAs~\cite{salem2018ml,shokri2017membership}. 
However, we find that dropout reduces $Adv_{Yeom}$ only by 40\% and $Adv_{Salem}$ by 42\% and weight decay may increase $Adv_{Salem}$ by, up to, 35\%.
Further, we also evaluate spatial dropout~\cite{tompson2015efficient}, as dropout is known to be ineffective for modern CNNs~\cite{wu2015towards}.
Spatial dropout is more consistent against MIAs than dropout: reducing $Adv_{Yeom}$ by 68\% and $Adv_{Salem}$ by 63\%.

Label smoothing is a recent technique that has seen widespread use~\cite{muller2019does}. 
We find that it reduces $Adv_{Yeom}$ by 43\%; however, it also consistently increases $Adv_{Salem}$; most notably by, up to, 59\% on CIFAR-100 in the $Max$ scenario.
Surprisingly, when the label smoothing reduces the generalization gap the most, it also increases the $Adv_{Salem}$ the most.
\citet{muller2019does} find that label smoothing erases information and encourages the model to treat incorrect classes as equally probable.
We hypothesize that models trained with label smoothing overfit on the smooth labels and produce more smooth output probabilities on $\mathcal{S}$ than on $\mathcal{D}$.
This helps Salem et al.'s attack as it relies on output probabilities; whereas, Yeom et al.'s attack cannot capitalize on it.
We believe this adverse effect poses a severe security risk to practitioners.

Further, we find that adding Gaussian noise is only effective if the task is simple enough to tolerate a noisy $\mathcal{S}$; however, as the task gets more complex, the noise hurts $Acc$ more than it reduces $Adv$.
Finally, even though early stopping is not the most effective, it is a consistent way of preventing MIAs; the earlier the training ends, the less successful both MIAs are, in exchange for reducing the accuracy.

\topic{Comparison with Differential Privacy}
To use as a baseline, we train Fashion-MNIST models using DP-SGD by \citet{abadi2016deep}.
Using DP on more complex tasks without hurting a model's accuracy requires additional techniques, such as transfer learning, that are not always available~\cite{abadi2016deep,rahman2018membership}.

We control the intensity of DP by varying the amount of noise added to the gradients, $\sigma_{DP}$, during training. 
The noise also determines the \emph{privacy budget}, $\varepsilon$; which is the formal limit on much information the model leaks.
We experiment with $0.01\!\leq\!\sigma_{DP}\!\leq\!10$, which results in $3\!\times 10^5\!>\!\varepsilon\!>\!0.15$.
Prior work considers budgets $\varepsilon\!<\!1$ formally acceptable for privacy~\cite{jayaraman2019evaluating}.

Within 5\% RAD, $\sigma_{DP}=0.1; \varepsilon \approx 3 \times 10^{3}$ reduces $Acc$ by 4.1\%, $Adv_{Yeom}$ by 82.3\% and $Adv_{Salem}$ by 100\%; falling behind techniques such as distillation, random cropping, spatial dropout, and dropout.
Within 15\% RAD, $\sigma_{DP}=0.25; \varepsilon \approx 2 \times 10^{2}$ reduces $Acc$ by 5.4\%, $Adv_{Yeom}$ by 90.1\% and $Adv_{Salem}$ by 100\%; again, falling behind distillation, random cropping, and weight decay.
Finally, even applying strong DP with $\sigma_{DP}=10; \varepsilon \approx 0.15$ falls behind regularization by reducing $Adv_{Yeom}$ by 92.7\%, while significantly hurting the $Acc$ by 16.7\%.

Our results show that, against known MIAs, regularization offers a more practical solution than DP.
Further, we confirm the prior findings that thwarting MIAs might not require strong formal guarantees~\cite{jayaraman2019evaluating}.
However, in Section~\ref{sec:comparison}, we show that certain mechanisms, such as distillation, might result in observably distinct behaviors for $\mathcal{S}$ and $\mathcal{D}$; even when the MIAs fail.
This indicates that the success against known MIAs might give a false sense of privacy.
DP, on the other hand, because of its worst-case guarantees, provides a true sense of privacy.

\topic{Takeaways}
Our experiments reveal that distillation, spatial dropout, and data augmentation with random cropping are consistent and effective mechanisms for defeating MIAs against CNN models.
We see that these mechanisms almost significantly prevent known MIAs while not preserving a model's accuracy.
Further, compared to DP, regularization reduces the attacker's success more with less utility penalty.
Moreover, defeating MIAs requires strong regularization; therefore, applying regularization to mainly increasing accuracy provides limited privacy benefits.
In terms of ease-of-tuning, early stopping is a practical choice as it yields predictable benefits without requiring any tuning.
Finally, label smoothing and, in certain cases, weight decay might exacerbate the vulnerability against MIAs that leverage output probabilities.
As a result, we believe these popular mechanisms might expose the practitioners to a serious privacy compromise.
%
%

\section{Combining Multiple Mechanisms}
\label{sec:combining}

In the previous section, we show that using regularization mechanisms in Table~\ref{table:reg_mechanisms} can prevent MIAs.
Especially for distillation, however, we observe diminishing returns as we increase the intensity of a single mechanism.
Here, we investigate whether combining two mechanisms has any benefits over individual mechanisms against MIAs.

\topic{Methodology}
We evaluate the effectiveness of the mechanism pairs on CIFAR-100 as the most vulnerable task to MIAs.
We set the mechanism hyperparameters in accordance with the RAD$<$5 scenario.
We use distillation ($T\!=\!50$) or random cropping ($c\!=\!16$) in each pair as they are individually the most effective mechanisms.
We use the results in the RAD$<$15 scenario to answer whether a pair, e.g., $T\!=\!50$ and $\lambda\!=\!1e\!-\!5$, is more effective than a single mechanism with higher intensity, e.g., $T\!=\!100$.


\begin{table}[htb]
\centering
\caption{Combining mechanisms on CIFAR-100. We pair each mechanism with distillation ($T$) or random cropping ($c$) in the RAD$<$5 scenario. Rows `\textbf{5}' and `\textbf{15}' present the results on individually applying $T$ and $c$, in the RAD$<$5 and RAD$<$15 scenarios, respectively.}

\label{table:combine_evaluation}
\adjustbox{max width=\linewidth}{
	\begin{tabular}{@{}c@{\hskip .09in}c@{\hskip .075in}c@{\hskip .075in}c@{\hskip .075in}|c@{\hskip .075in}c@{\hskip .075in}c@{}}
		\toprule
		\multirow{2}{*}{} & \multicolumn{3}{c|}{$T=50$} & \multicolumn{3}{c}{$c = 16$} \\ 
		\textbf{} & $Adv_{Yeom}$ & $Adv_{Salem}$ & $Acc$ & $Adv_{Yeom}$ & $Adv_{Salem}$ & $Acc $ \\ \midrule \midrule
		\textbf{5} & -68.6\% & -99.4\% & -4.2\% & -79.8\% & -87.6\% & -4.4\% \\
		\textbf{15} & -68.8\% & -100\% & -6.1\% & -86.9\% & -100.0\% & -5.9\% \\ \hline \addlinespace[0.1cm]
		$T$ & - & - & - & -93.1\% & -100.0\% & -10.1\% \\
		$\alpha$ & -68.5\% & -100.0\% & -2.5\% & -87.1\% & -100.0\% & -10.5\% \\
		$\sigma$ & -69.1\% & -100.0\% & -4.0\% & -82.3\% & -88.5\% & -5.3\% \\
		$c$ & -93.1\% & -100.0\% & -10.1\% & - & - & - \\
		$p$ & -72.1\% & -100.0\% & -6.2\% & -85.0\% & -89.4\% & -10.3\% \\
		$q$ & -89.8\% & -100.0\% & -10.0\% & -96.0\% & -100.0\% & -31.4\% \\
		$\lambda$ & -96.5\% & -100.0\% & -48.8\% & -84.7\% & -88.1\% & -5.8\% \\
		$E$ & -71.0\% & -100.0\% & -4.4\% & -92.3\% & -95.3\% & -24.9\% \\ \bottomrule
	\end{tabular}
}
\end{table}

\topic{Evaluating the Combinations}
Table~\ref{table:combine_evaluation} summarizes the effectiveness of combining two mechanisms.
For distillation, between $T\!=\!50$ and $T\!=\!100$, we see diminishing returns in resilience against MIAs; whereas, for cropping, between $c\!=\!16$ and $c\!=\!20$, the improvement is more significant.

We observe that combining distillation with other mechanisms counters MIAs more than moving from $T\!=\!50$ to $T\!=\!100$. 
For $T\!=\!100$, the $Adv_{Yeom}$ and $Adv_{Salem}$ decrease by 69\% and 100\%, respectively; however, almost all the combinations of distillation, decrease $Adv_{Yeom}$ more.
The combinations with label smoothing and Gaussian noise mechanisms lead to less drop in $Acc$, while achieving comparable resilience.
Combining with weight decay, even though reducing $Adv_{Yeom}$ by 96\%; causes a significant 49\% accuracy drop.
Combining with spatial dropout or cropping, achieves an impressive $\sim$90\% drop in $Adv_{Yeom}$; while still staying within RAD$<$15.

On the other hand, random cropping does not benefit from combinations as much as distillation. 
For $c\!=\!20$, the $Adv_{Yeom}$ and $Adv_{Salem}$ decrease by 87\% and 100\%, respectively; however, almost all combinations with cropping show similar resilience with more accuracy drop.
Most notably, the combination with spatial dropout leads to 96\% reduction in $Adv_{Yeom}$, while causing 31\% drop in $Acc$.
Across the board, this setting achieves the highest resilience while causing significantly less accuracy drop than the combination of distillation and weight decay.

Finally, combining early stopping with random cropping hurts the accuracy significantly; whereas, with distillation, it results in marginal changes.
The reason is that distillation speeds up the training~\cite{hinton2015distilling}; while aggressive random cropping slows it down by randomly removing features.
Compared to distillation, the models trained with random cropping require more passes over the training set to learn the important features.
To validate, we first combine $T\!=\!50$ with $E\!=\!\{20, 15, 10, 5\}$ and see 3\%, 4\%, 6\%, and 8\% accuracy drops, respectively. 
On the other hand, when we combine $c\!=\!16$ with $E\!=\!\{20, 15, 10, 5\}$, it results in 14\%, 25\%, 29\%, and 46\% 
utility loss, respectively. 
This implies that combining two mechanisms that have similar effects on a model might counteract with each other and lead to a substantial accuracy drop.

\section{Exposing the False Sense of Privacy}
\label{sec:comparison}

This section aims at designing a white-box metric for providing more realistic lower bounds for the vulnerability to MIAs than existing black-box attacks.
In the previous sections, we find that some regularization mechanisms significantly reduce the attack's success while preserving the accuracy.
This might convince the practitioners to use these mechanisms to safeguard the privacy of their training data.
Here, in the absence of formal guarantees, we apply our metric to study whether this sense of privacy is well placed.

Our \emph{distance-to-confident} (DtC) metric builds on the intuition that the samples in $\mathcal{S}$ are closer to a model's confident decision regions the ones in $\mathcal{D}$.
In these regions, a model's predictions are highly confident and produce low losses.
When a model overfits on $\mathcal{S}$, it places the majority of its training samples in these regions; which enables the MIAs to discern them.
We expect that, for the samples outside these regions, the distance to a confident region is less for the samples in $\mathcal{S}$ than the ones in $\mathcal{D}$.
This discrepancy compromises privacy as it provides an adversary with leverage to characterize the training samples.

\topic{The Distance-to-Confident (DtC) Metric}
We specify a sample $(x,y)$ as not in a confident region of the model $F$, if $\mathcal{L}(F(x), y) > \tilde{\mathcal{L}}$, where $\tilde{\mathcal{L}}$ is the $F$'s median loss on its $\mathcal{S}$.
The basis of our metric is finding the shortest distance from $x$ to a confident region.
Since it is not possible to directly measure the distances in the decisions of a deep learning model, prior work has used adversarial input perturbations as a proxy~\cite{tramer2017space}.
Similarly, we use PGD perturbations~\cite{madry2017towards} to measure the distance from a sample $(x,y)$ to a confident region.
As PGD perturbations require access to a model's all internal details, such as its gradients, DtC is a white-box metric.
However, the practical ways of applying adversarial perturbations in the black-box setting~\cite{pmlr-v97-guo19a} makes using our DtC metric towards a powerful membership inference attack possible.
Here, we utilize it as a defensive metric and leave this conversion into an attack as future work.

To move $x$ to a confident region, we start from $x^0 = x$ and, at each step, we perturb $x^{t}$ as $x^{t+1}$ to reduce the $F$'s loss, $\mathcal{L}_t = \mathcal{L}(F(x^{t}), y)$.
These perturbations slowly move $x$ to a confident, low-loss, region and we stop perturbing when $\mathcal{L}_t < \tilde{\mathcal{L}}$.
The DtC of $x$ for the model $F$ is then given by the distance from $x$ to the final $x^t$, i.e., $\mathcal{C}_F(x) = ||x^t - x||_\infty$.

In our experiments, we apply PGD for 100 steps and set the step size, $||x^{t+1} - x^{t}||_\infty$, as $0.001$.
This results in $0.001 \leq \mathcal{C}_F(x) \leq 0.1$.
Before reporting, we multiply the DtC scores by 1000 to bring them between 1 and 100.

\topic{Measuring DtC for Baseline Models}
Table~\ref{table:baseline_stab} presents the average DtC scores, $\overline{\mathcal{C}}$, of our baseline models.
We define a model's \emph{DtC gap}, $\mathcal{C_{D - S}}$, as its relative difference between $\overline{\mathcal{C}}_\mathcal{D}$ and $\overline{\mathcal{C}}_\mathcal{S}$.
The larger DtC gaps indicate greater levels of vulnerability and may enable an adversary to more accurately differentiate between $\mathcal{D}$ and $\mathcal{S}$.

First, we see that the average DtC scores are significantly lower on $\mathcal{S}$ than on $\mathcal{D}$; confirming our initial intuition.
The DtC gap also correlates well with the success of existing MIAs.
Further, as the task complexity goes up, the DtC scores on $\mathcal{S}$ decrease, as a result of more severe overfitting.
These findings indicate that the DtC is a reasonable metric for studying the vulnerability against MIAs.

\begin{table}[htb]
    \caption{Average DtC scores of the baseline models on their training set, $\mathcal{S}$, and their test set, $\mathcal{D}$. We also present the DtC gap, $\mathcal{C_{S - D}}$, as a percentage.}
    \label{table:baseline_stab}
    \centering
    \begin{small}
    \begin{sc}
    \begin{tabular}{@{}lcccc}
        \toprule
        \textbf{Task} &\textbf{$\overline{\mathcal{C}}_\mathcal{D}$} &\textbf{$\overline{\mathcal{C}}_\mathcal{S}$} & $\mathcal{C_{D - S}}$ & \textbf{$Adv_{Yeom}$}\\
        \midrule
        \small{F-MNIST}&\small{5.2}&\small{2.8}&\small{46.1\%}&\small{9.6\%}\\
        \small{CIFAR-10}&\small{2.2}&\small{1.0}&\small{54.5\%}&\small{35.8\%}\\
        \small{CIFAR-100}&\small{8.0}&\small{1.1}&\small{86.2\%}&\small{68.9\%}\\
        \bottomrule
    \end{tabular}
    \end{sc}
    \end{small}
\end{table}
\raggedbottom

\topic{Measuring DtC for Regularized Models}
Table~\ref{table:stab_reg} presents the average DtC scores of the models with regularization, in the RAD$<$15\% scenario.
First note that, even when it fully eliminates the MIAs, random cropping on Fashion-MNIST still displays a 5\% DtC gap; indicating the vulnerability might still persist.
Even though it significantly reduces the MIA success, models trained with distillation have usually large $\mathcal{C_{D - S}}$.
For example, on CIFAR-10, distillation achieves 5\% $Adv_{Yeom}$ and still displays 22\% stability gap; whereas weight decay achieves a smaller 20\% gap with 7\% $Adv_{Yeom}$.
Similarly, on CIFAR-100, distillation achieves 22\% $Adv_{Yeom}$ and 64\% gap; again larger than 60\% gap spatial dropout achieves.
This hints that distillation's success against MIAs might be giving a false sense of privacy as the models' DtC gap is larger than expected.

Further, we see that adding Gaussian noise also leads to large DtC gaps, even though it prevents the MIAs.
For example, on Fashion-MNIST, it leads to a 12\% gap, while achieving only 1\% $Adv_{Yeom}$.
We hypothesize that adding noise does not prevent overfitting but it merely obfuscates where the model's confident regions are.
This reduces the success of the known MIAs; however, our DtC metric finds the obfuscated regions and exposes the vulnerability.

For dropout, spatial dropout and random cropping, we observe that their success against MIAs aligns well with their DtC gaps.
This implies that these mechanisms are not misleading and provide more reliable privacy.

Finally, we measure the DtC gap for the models that combine multiple mechanisms in Section~\ref{sec:combining}. 
On CIFAR-100, combining cropping ($c\!=\!16$) with distillation ($T\!=\!50$) achieves 5\% $Adv_{Yeom}$; however, with 22\% DtC gap.
Similarly, the combination of cropping and spatial dropout ($q\!=\!0.25$) results in 3\% $Adv_{Yeom}$ with 9\% DtC gap.
Cropping with early stopping ($E\!=\!5$) shrinks the DtC gap to $\sim$0\% and achieves 2\% $Adv_{Yeom}$; however, it also causes 46\% accuracy drop. 
Overall, combining can improve the resilience against known MIAs; however, it cannot prevent the DtC gap and the give a false sense of privacy.


\begin{table}[htb]
	\caption{Average DtC scores of the models with regularization in the RAD$<$15\% scenario. Each cell contains the model's $\overline{\mathcal{C}}_\mathcal{D}$, $\overline{\mathcal{C}}_\mathcal{S}$, $\mathcal{C_{D - S}}$ and $Adv_{Yeom}$; in its top left, top right, bottom left and bottom right quadrants, respectively.}
	\label{table:stab_reg}
	\centering
	\adjustbox{max width=\linewidth}{%
		\begin{tabular}{@{}l@{}c@{}c@{}c@{}}
			\toprule
			\textbf{R.} & \textbf{F-MNIST} & \textbf{CIFAR-10}  & \textbf{CIFAR-100}\\
			\midrule
			$T$ & \splitcell{20.4}{19.0}{6.9\%}{0.5\%}  &  \splitcellc{4.5}{3.5}{22.2\%}{5.4\%} & \splitcellr{6.7}{2.4}{64.2\%}{21.6\%}\\ \addlinespace[0.14cm]
			
			$\alpha$ & \splitcell{56.9}{56.4}{0.9\%}{0.4\%} & \splitcellc{4.3}{2.7}{37.2\%}{19.3\%} & \splitcellr{13.3}{1.9}{85.7\%}{43.8\%}\\ \addlinespace[0.14cm]
			
			$\sigma$ & \splitcell{10.6}{9.3}{12.3\%}{1.2\%} & \splitcellc{3.6}{1.6}{55.6\%}{19.7\%} & \splitcellr{7.3}{1.3}{82.2\%}{57.4\%}\\ \addlinespace[0.14cm]
			
			$c$ & \splitcell{19.9}{18.9}{5.0\%}{0.0\%} & \splitcellc{3.9}{3.7}{3.9\%}{2.2\%} & \splitcellr{5.2}{4.2}{19.2\%}{9.0\%}\\ \addlinespace[0.14cm]
			
			$p$ & \splitcell{44.2}{41.5}{6.1\%}{1.4\%} & \splitcellc{20.6}{18.6}{9.7\%}{13.1\%} & \splitcellr{5.5}{1.2}{78.2\%}{57.1\%}\\ \addlinespace[0.14cm]
			
			$q$ & \splitcell{67.5}{65.3}{3.3\%}{1.0\%} & \splitcellc{10.2}{9.1}{10.8\%}{2.8\%} & \splitcellr{13.7}{5.5}{59.8\%}{32.4\%}\\ \addlinespace[0.14cm]
			
			$\lambda$ & \splitcell{54.2}{53.8}{1.1\%}{0.6\%} & \splitcellc{4.5}{3.6}{20.0\%}{6.7\%} & \splitcellr{8.0}{1.1}{86.2\%}{55.9\%}\\ \addlinespace[0.14cm]
			
			$E$ & \splitcell{37.3}{34.9}{6.4\%}{1.9\%} & \splitcellc{6.4}{5.4}{15.6\%}{5.7\%} & \splitcellr{7.8}{2.3}{70.5\%}{46.1\%}\\
			\bottomrule
		\end{tabular}
	}
\end{table}
\raggedbottom

\topic{Measuring DtC for Differentially-Private Models}
Models trained with DP provide worst-case privacy guarantees, regardless of the attack. 
As a result, given strong enough guarantees, we expect DP not to display any significant DtC gap.
To confirm, we analyze the models trained with increasing intensities of DP, $0.01\!\leq\!\sigma_{DP}\!\leq 10;\ 3\!\times\!10^5\!>\!\varepsilon\!>\!0.15$, on Fashion-MNIST.

\begin{figure}[t]
	\centering
	\includegraphics[width=\linewidth]{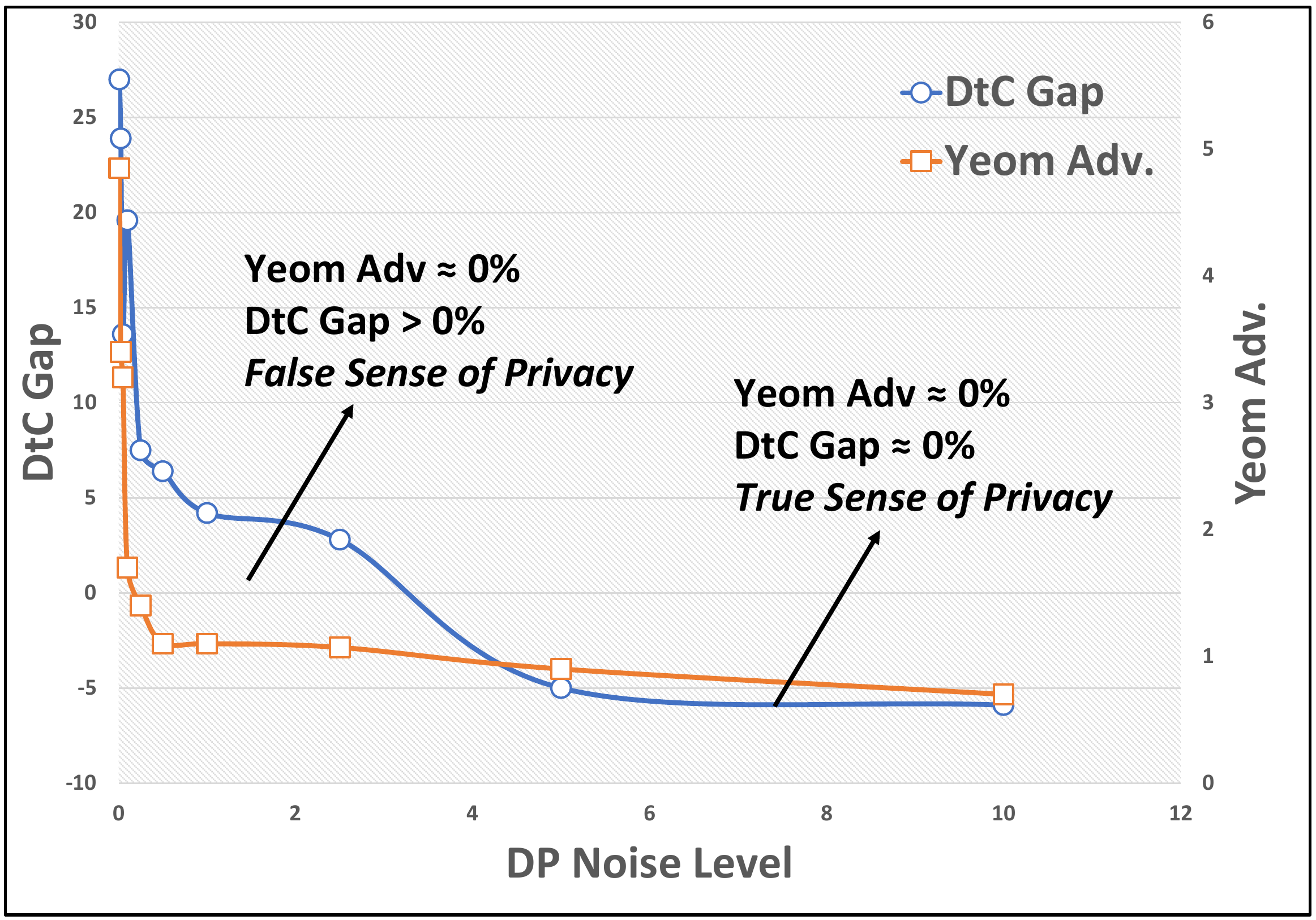}
	\caption{Measuring the DtC gap for models trained with different levels of differential privacy.}
	\label{fig:dp_dtc}
\end{figure}

Figure~\ref{fig:dp_dtc} shows how the DtC gap and the success against MIAs change, along with stronger DP guarantees.
We see that even when DP achieves $Adv_{Yeom} \approx 1\%$ with $0.25\!\leq\!\sigma_{DP}\!\leq 1$; the large DtC Gap indicates the vulnerability still persists.
The privacy budgets in these settings, $2\!\times\!10^{2}\!>\varepsilon\!>\!2$, are usually considered to be insufficient.

On the other hand, for stronger privacy guarantees, $2.5\!\leq\!\sigma_{DP}\!\leq\!10;\ 2\!>\!\varepsilon\!>\!0.15$, the DtC gap shrinks; even though, $Adv_{Yeom}$ still stays around $\sim$1\%.
This highlights the utility of having stronger DP guarantees, even though there is no practical attack that can achieve the worst-case.
We believe our DtC metric also provides practitioners a more reliable way of adjusting DP's privacy budget, as we show that success against MIAs underestimates the vulnerability and DP's guarantees might be too restrictive.

%
%

\section{Conclusions}
\label{sec:conclusions} 
We conduct a systematical study on the effectiveness of regularization in mitigating membership inference attacks (MIAs) against deep learning models.
We identify the effective, e.g., random cropping, and ineffective, e.g., weight decay, mechanisms, which allows us to propose practical guidelines.
Further, we reveal that the label smoothing mechanism, by improving both the generalization and the attack performance, challenges the notion that MIAs stem from overfitting.
Our findings on combining multiple mechanisms highlight an important research direction of finding optimal combinations to prevent MIAs.
Finally, we design a white-box metric to provide realistic lower bounds for the vulnerability.
Our metric helps us to show that, even when existing MIAs fail, a model may still remain vulnerable to unknown future attacks.
This exposes the false sense of privacy given by certain regularization mechanisms, e.g., distillation; and also illustrates the usefulness of differential privacy that prior work considers as restrictive in practice.
We hope that our work acts as a bridge between the practical and formal solutions to the privacy leakage problem.

\newpage
\section*{Acknowledgements}
\label{sec:ack}
This research was partially supported by the Department of Defense.

\bibliographystyle{icmlrefs}
\bibliography{refs}


\end{document}